\documentclass[10pt,twocolumn,letterpaper]{article}

\usepackage{cvpr}              

\usepackage{amsmath,amsfonts}
\usepackage[linesnumbered,ruled]{algorithm2e}
\usepackage{textcomp}
\usepackage{stfloats}
\usepackage{verbatim}
\usepackage{graphicx}
\usepackage{multirow}
\usepackage{blindtext}
\usepackage{xcolor}
\usepackage{booktabs}
\usepackage{xcolor,colortbl}
\usepackage[accsupp]{axessibility}
\definecolor{cvprblue}{rgb}{0.21,0.49,0.74}

\usepackage[pagebackref,breaklinks,colorlinks,citecolor=cvprblue]{hyperref}

\def\paperID{16014} %
\def\confName{CVPR}
\def\confYear{2024}

\title{MTLoRA: A Low-Rank Adaptation Approach for Efficient Multi-Task Learning}

\author{Ahmed Agiza\thanks{\noindent The first two authors contributed equally to this work.}\\
Brown University\\
Providence, RI\\
{\tt\small ahmed\_agiza@brown.edu}
\and
Marina Neseem\footnotemark[1]\\
Brown University\\
Providence, RI\\
{\tt\small marina\_neseem@brown.edu}
\and
Sherief Reda\\
Brown University\\
Providence, RI\\
{\tt\small sherief\_reda@brown.edu}
}

\begin{document}

\maketitle

\begin{abstract}

Adapting models pre-trained on large-scale datasets to a variety of downstream tasks is a common strategy in deep learning. 
Consequently, parameter-efficient fine-tuning methods have emerged as a promising way to adapt pre-trained models to different tasks while training only a minimal number of parameters. 
While most of these methods are designed for single-task adaptation, parameter-efficient training in Multi-Task Learning (MTL) architectures is still unexplored. 
In this paper, we introduce \textit{MTLoRA}, a novel framework for parameter-efficient training of MTL models. 
\textit{MTLoRA} employs \textit{Task-Agnostic} and \textit{Task-Specific} Low-Rank Adaptation modules, which effectively disentangle the parameter space in MTL fine-tuning, thereby enabling the model to adeptly handle both task specialization and interaction within MTL contexts. 
We applied MTLoRA to hierarchical-transformer-based MTL architectures, adapting them to multiple downstream dense prediction tasks. 
Our extensive experiments on the PASCAL dataset show that MTLoRA achieves higher accuracy on downstream tasks compared to fully fine-tuning the MTL model while reducing the number of trainable parameters by $3.6\times$. 
Furthermore, \textit{MTLoRA} establishes a Pareto-optimal trade-off between the number of trainable parameters and the accuracy of the downstream tasks, outperforming current state-of-the-art parameter-efficient training methods in both accuracy and efficiency. Our code is publicly available.\footnote{https://github.com/scale-lab/MTLoRA.git}

\vspace{-10pt}
\end{abstract}

\section{\textbf{Introduction}}
\label{sec:intro}
\begin{figure}[t]
  \centering
\includegraphics[width=0.95\linewidth]{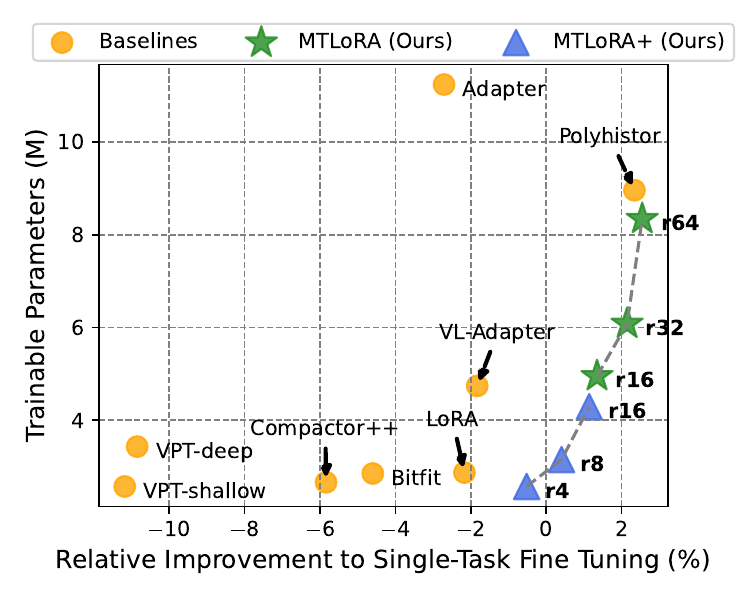}
\vspace{-6pt}
\caption{\textit{MTLoRA} versus state-of-the-art parameter-efficient training approaches using Swin-Tiny vision transformer as a backbone. \textbf{r} represents the different ranks for the low-rank decomposition modules inside MTLoRA. }
\vspace{-20pt}
\label{fig:mtlora_vs_sota}
\end{figure}

General-purpose vision and language models, particularly those trained on large-scale datasets, show remarkable adaptability to a wide range of downstream tasks \cite{liu2021swin, touvron2023llama}.
However, individually fine-tuning all parameters of these models for every downstream task poses significant efficiency challenges.
This approach becomes increasingly inefficient as the number of tasks grows, especially in environments constrained by computational resources. 

Therefore, there is a need to develop resource-efficient fine-tuning techniques \cite{hu2021lora,mahabadi2021parameter,liu2022polyhistor,hu2023llm}.
These methods aim to optimize training efficiency by limiting the number of trainable parameters, all while attempting to preserve or enhance task-specific fine-tuning.
Most existing parameter-efficient adaptation methods are primarily tailored for single-task adaptation, and they may lose their effectiveness when applied to multi-task learning (MTL) scenarios.
This is attributed to the inherent complexity of MTL, where the goal is to optimize the performance of a single model across a spectrum of tasks, introducing an additional layer of complexity. 
Moreover, focusing solely on individual task adaptation overlooks the potential benefits of cross-task knowledge sharing. 
Such knowledge sharing in an MTL context can significantly enhance the performance of each task \cite{crawshaw2020multi,zhang2021survey}.

\begin{figure}[t]
  \centering
  \begin{subfigure}{0.8\linewidth}
    \includegraphics[width=\linewidth]{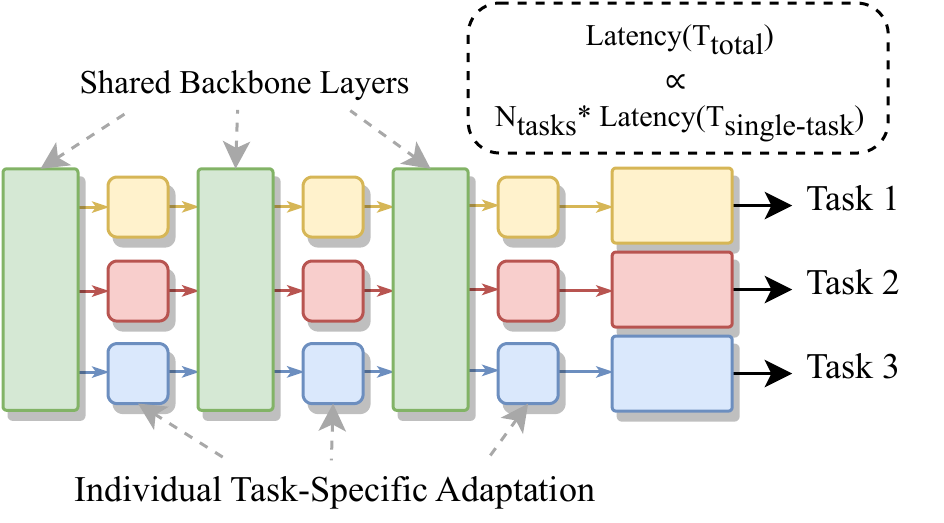}
    \caption{Individual Task-Specific Adaptation}
    \label{fig:individual_task_specific Adaptation}
  \end{subfigure}
    \\
  \begin{subfigure}{0.8\linewidth}
    \includegraphics[width=\linewidth]{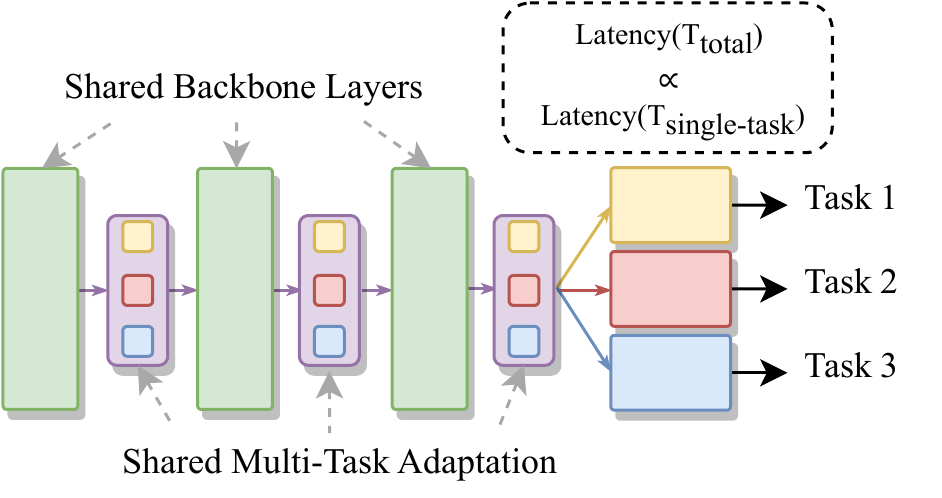}
    \caption{Shared Multi-Task Adaptation (Ours)}
    \label{fig:shared_multi_task_adaptation}
  \end{subfigure}
  \caption{\textit{(a) Individual Task Adaptation} results in parallel execution paths for each task, resulting in inference and training time that scales linearly with the number of tasks. 
  On the other hand, (b) \textit{Shared Multi-Task Adaptation} maintains inference and training time close to the single task model since only the decoders are executed separately. 
  }
  \label{fig:mtltypes}
  \vspace{-20pt}
\end{figure}

To realize multi-task adaptation using existing methods \cite{liu2022polyhistor, mahabadi2021parameter}, individual modules specific to the different tasks have to be added and adapted to one downstream task at a time as shown in Figure \ref{fig:individual_task_specific Adaptation}.
This approach enables customization and improvement for each task's unique needs, especially useful when tasks have unique characteristics or require specialized knowledge \cite{ruder2019latent,misra2016cross}.
However, this strategy incurs a significant drawback in terms of efficiency during both training and inference. 
Due to the task-specific nature of the fine-tuning, the model must be trained and inferred separately for each task. 
This results in a proportional increase in computational cost and time with the number of tasks. 
For instance, adapting a model to five distinct tasks would require five separate training passes. 
Similarly, during inference, the backbone needs to be executed five times, once for each task, leading to a linear escalation in inference and training duration as the task count increases.

Our research diverges from conventional parameter-efficient adaptation methods by concentrating on parameter-efficient training specifically for multi-task learning (MTL) architectures. 
In MTL models, a single shared backbone is trained to simultaneously extract feature representations for various downstream tasks \cite{dai2016instance,liu2019end,xu2018pad}. 
MTL offers significant efficiency advantages since the shared backbone is executed only once, as shown in Figure \ref{fig:shared_multi_task_adaptation}, leading to more resource-efficient training and inference processes where latency does not increase linearly with the number of tasks.
Despite these advantages, the aspect of parameter-efficient training in MTL architectures remains largely unexplored. 

The primary challenge in fine-tuning MTL models efficiently lies in addressing task conflicts during fine-tuning. 
These conflicts arise when different tasks have competing demands or induce divergent updates in the model. Hence, the focal point for many MTL architectures \cite{javaloy2021rotograd,liu2021conflict} is to balance these conflicting updates. 
Consequently, a pivotal question emerges: \textit{How can we efficiently adapt a single shared backbone to serve multiple tasks without sacrificing the individual performance of each task?}

In pursuit of this objective, we introduce \textit{MTLoRA} \-- a novel framework designed for parameter-efficient fine-tuning of MTL models.
\textit{MTLoRA} addresses the challenges of fine-tuning a shared backbone to effectively serve multiple downstream tasks, particularly under the constraints of conflicting task requirements.
This is accomplished through a strategic combination of \textit{Task-Agnostic} and \textit{Task-Specific} low-rank decomposition modules. 
By fine-tuning these modules, \textit{MTLoRA} successfully untangles the parameter space involved in MTL fine-tuning, enabling the model to balance between learning shared features and those specific to individual tasks. 
Remarkably, \textit{MTLoRA} demonstrates superior accuracy in downstream tasks compared to fully fine-tuning the entire MTL model while requiring the training of significantly fewer parameters. 
This enhanced performance is attributed to \textit{MTLoRA}'s ability to facilitate positive knowledge sharing during fine-tuning, thereby improving the effectiveness of learning each downstream task.
\textbf{Our contributions} can be summarized as follows:
\begin{itemize}
\item To the best of our knowledge, \textit{MTLoRA} is the first to address the problem of parameter-efficient training of multi-task learning models.
\textit{MTLoRA} effectively balances between learning both shared and task-specific features during parameter-efficient fine-tuning.
\item We design novel \textit{Task-Agnostic} and \textit{Task-Specific} low-rank adaptation modules leveraging them to adapt a shared vision-transformer backbone to multiple downstream dense prediction tasks.
\item We observe that adding low-rank adaptation to the patch-merging layers in vision transformers, a practice not previously explored, significantly improves the accuracy-efficiency trade-off during fine-tuning MTL models. We highlight that observation by introducing \textit{MTLoRA+}.
\item We apply \textit{MTLoRA} and \textit{MTLoRA+} to a hierarchical-transformer-based MTL architecture.
\textit{MTLoRA} demonstrates superior accuracy in downstream tasks compared to fully fine-tuning the entire MTL model while training significantly less number of parameters.
In addition, \textit{MTLoRA} dominates state-of-the-art parameter-efficient training approaches, as shown in Figure ~\ref{fig:mtlora_vs_sota}. 
\end{itemize}

The rest of the paper is organized as follows. We review the related work in Section~\ref{sec:related-work}. Then, we introduce our MTLoRA framework in Section ~\ref{sec:methodology}. Next, we show the setup and evaluation of MTLoRA in Section~\ref{sec:expr}. Finally, we conclude in Section~\ref{sec:conclusion}.

\section{\textbf{Related Work}}
\label{sec:related-work}
\textbf{Multi-task Learning:}
Multi-task learning is commonly used to learn various related tasks simultaneously \cite{misra2016cross, astmt, neseem2023adamtl}. The typical design of a multi-task architecture includes an encoder to distill feature representations from input frames and a set of task-specific decoders for generating predictions unique to each downstream task \cite{zhang2021survey}. 
An important aspect to consider within these multi-task architectures is the mechanism of information sharing. The two main strategies are soft sharing and hard sharing. 
Soft parameter sharing involves each task having its own set of backbone parameters, with the primary objective of facilitating cross-task information exchange. 
On the other hand, hard parameter sharing employs a shared set of parameters within the backbone, with each task employing independent decoders for output generation \cite{kendall2018multi,sener2018multi,bekoulis2018adversarial}.
Further classification of these architectures takes into account the stage at which task interactions occur - leading to the categorization into encoder-focused and decoder-focused frameworks \cite{xu2018pad}. Encoder-focused architectures centralize information exchange within the encoder stage \cite{vandenhende2020mti,mtan}, whereas, in decoder-focused architectures, tasks exchange information during the decoding stage. Notably, some models adopt a more integrative approach, allowing for cross-task information sharing to occur at encoder and decoder stages \cite{misra2016cross}.

\textbf{Parameter-Efficient Training for Single-Task Models:}
Parameter-efficient training (PEFT) has become increasingly important, especially when dealing with large-scale pre-trained models \cite{hu2021lora,zhang2023llama,gao2023llama,hu2023llm} since traditional fine-tuning methods, which involve adjusting a significant portion of a model's parameters for specific tasks, can be resource-intensive. 
Two common techniques in this domain are adapters \cite{zhang2023llama,gao2023llama} and Low-Rank Adaptation (LoRA) \cite{hu2021lora, dettmers2023qlora}.
\textit{Adapters} are lightweight modules inserted between the layers of a pre-trained model, which allows for targeted modifications to the model's behavior without altering the original pre-trained weights.
This approach is beneficial as it reduces the number of parameters that need to be fine-tuned, thus lowering the computational burden. Adapters have shown effectiveness in various tasks, providing a flexible and efficient way to adapt large models to specific tasks or datasets.
However, one limitation of adapters is the additional parameters they introduce, which can lead to increased computational requirements during inference.
On the other hand, \textit{LoRA} offers a different approach to PEFT.
LoRA involves modifying the weight matrices of a pre-trained model using low-rank decomposition. This method allows for fine-tuning the model's behavior while maintaining the original structure and size of the weight matrices.
The key advantage of LoRA is that it does not introduce additional parameters during the model's runtime. Instead, it updates the pre-existing weights to enhance the model's performance on new tasks with minimal increase in computational requirements.
LoRA has been successfully applied in various fields, including NLP \cite{hu2021lora,dettmers2023qlora,chavan2023one,chen2023longlora} and computer vision \cite{he2023parameter}, demonstrating its versatility and effectiveness.
However, these methods, while efficient, only focus on single-task models. 

\textbf{Parameter-Efficient Training for Multi-Task Models:}
In a multi-task setting, PEFT is more challenging as the model must cater to the needs of multiple tasks simultaneously, often leading to increased complexity and potential for task interference. Consequently, some recent studies have proposed new solutions to extend the benefits of PEFT for multi-task adaptation.
One such approach is the Hypernetworks \cite{mahabadi2021parameter}, which uses shared networks to generate adapter parameters for all layers conditioned on the task, thus allowing for the sharing of information across different tasks while enabling task-specific adaptation through task-specific adapters. Building on top of it, Polyhistor \cite{liu2022polyhistor} explores PEFT in the domain of dense vision tasks, specifically on hierarchical vision transformers. Polyhistor proposes two ideas: decomposing hypernetworks into low-rank matrices and using custom kernels to scale fine-tuning parameters to the different transformer blocks.
However, these two approaches rely on separate execution of the model for each task to apply its adapter, which does not benefit from the MTL's potential for efficient training or inference.

\section{\textbf{Methodology}}
\label{sec:methodology}
\textbf{Problem Setting:}
Given a general-purpose transformer-based backbone pre-trained on large-scale image datasets
(e.g., ImageNet \cite{deng2009imagenet}), our goal is to efficiently adapt it to several downstream tasks in a Multi-Task Learning (MTL) architecture setting.
We are considering the common MTL architecture with one shared encoder and multiple task-specific decoders, as shown in Figure \ref{fig:shared_multi_task_adaptation}.
Following the existing works in parameter-efficient training, the criteria of parameter-efficient MTL training include the accuracy of downstream tasks and the number of training parameters.

\noindent \textbf{Method Overview:}
Our approach for efficiently adapting MTL models to various downstream tasks consists of two novel aspects: (1) efficiently share homogeneous information across tasks via a pool of task-agnostic and task-specific low-rank matrices and
(2) efficiently allow multi-scale task-specific feature sharing between the shared-encoder and task-specific decoders of the MTL architecture. 

This section is organized as follows.
We start with an overview of the used MTL architecture in Subsection \ref{baseline_mtl_arch}.
Then, we propose our parameter-efficient task-specific adaptation method in Subsection \ref{mtlora}.
In Subsection \ref{multi_scale_feature_sharing}, we propose our multi-scale task-specific efficient feature-sharing method.
Finally, we explore the effect of fine-tuning the non-attention modules in Subsection \ref{training}.

\subsection{MTL Architecture Overview}
\label{baseline_mtl_arch}

We first establish a Multi-Task Learning (MTL) framework designed for experimentation with parameter-efficient adaptation in MTL contexts. 
Aligning with established MTL methodologies \cite{ye2022inverted, vandenhende2020mti}, our model comprises three main components: a shared hierarchical encoder, task-specific inter-scale fusion modules, and a pool of task-specific decoders. 
We adopt an off-the-shelf hierarchical vision transformer as the shared encoder \cite{liu2021swin}, which extracts visual features from input frames for all downstream tasks. 
The hierarchical structure of the encoder allows for capturing visual features at various scales, providing a comprehensive representation of the input data.
The extracted multi-scale visual features are then fused and processed by various task-specific decoders to execute the downstream tasks.
Our MTL framework is designed to accommodate different vision transformer and decoder architectures, which makes our parameter-efficient adaptation approach suitable for a wide range of MTL architectures.

To effectively adapt the MTL architecture to various downstream tasks, we draw inspiration from the low-rank adaptation (LoRA) technique commonly employed in Language Models \cite{hu2021lora}, traditionally used for single-task adaptation. 
Our primary inquiry is: \textit{'How does fine-tuning low-rank matrices perform when optimizing for multiple visual downstream tasks?'}.
Previous studies in low-rank adaptation mainly aim to identify a unique set of low-rank matrices for adapting the encoder to an individual downstream task \cite{jia2022visual, sung2022vl, mahabadi2021parameter}, as illustrated in Figure \ref{fig:individual_task_specific Adaptation}. 
This approach requires running the entire model separately for each of the tasks, which is inefficient for real-time applications. 
In contrast, our research seeks to develop a single set of low-rank adaptation matrices that are applicable across multiple downstream tasks, as depicted in Figure \ref{fig:shared_multi_task_adaptation}. 
This methodology enables a single execution of the backbone for all tasks.
Multi-task learning often presents challenges, such as the “conflicting gradients problem” \cite{javaloy2021rotograd}. 
This issue becomes more pronounced in low-rank adaptation due to the limited number of trainable parameters.

\begin{figure*}
  \centering
  \begin{subfigure}{0.61\linewidth}
    \includegraphics[width=0.85\linewidth]{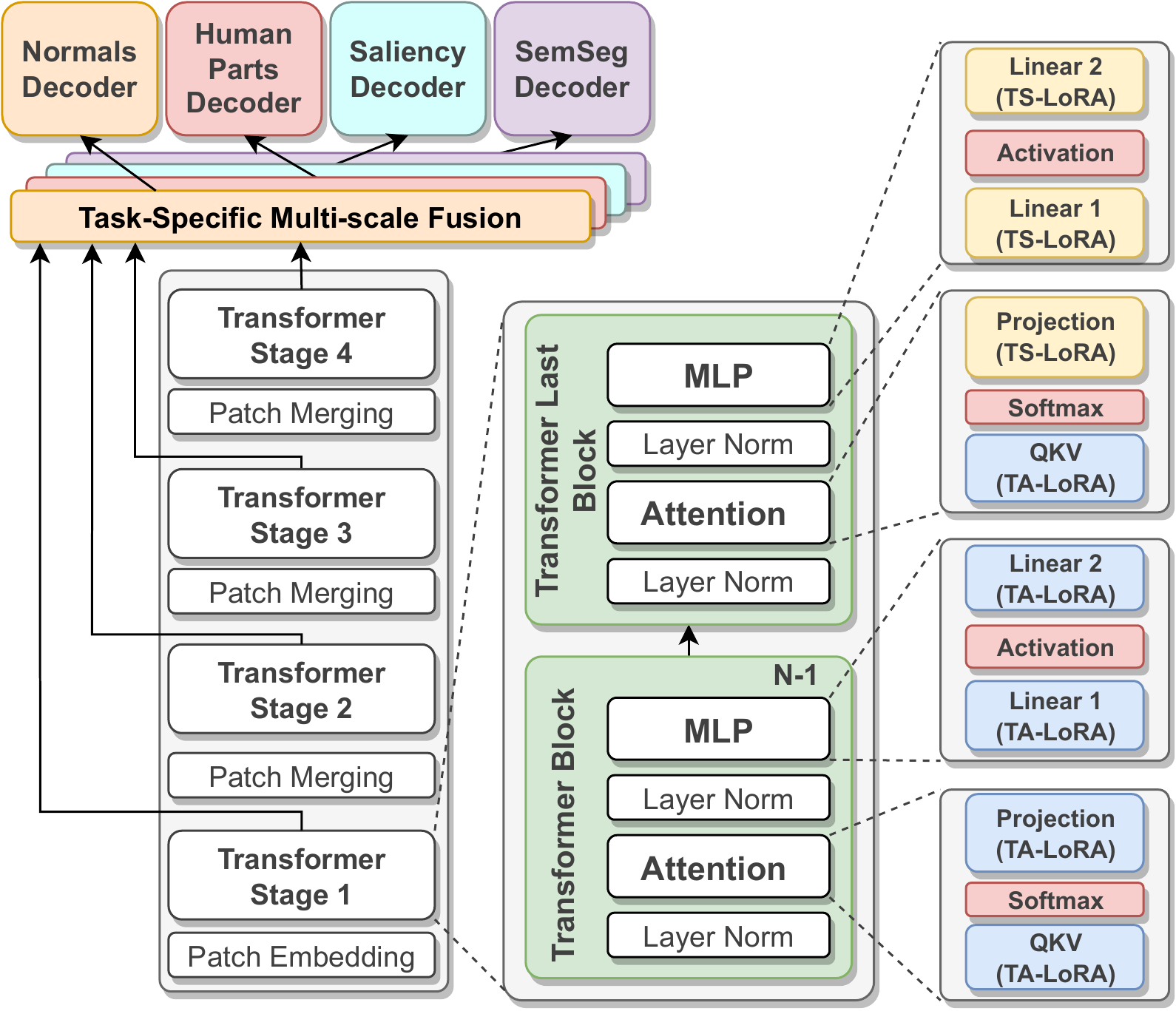}    
    \caption{Low-Rank Adaptation matrices are applied to all the blocks of the shared backbone.}
    \label{fig:backboneMtlora}
  \end{subfigure} 
  \hfill
  \begin{subfigure}{0.38\linewidth}
   \includegraphics[width=0.5\linewidth]{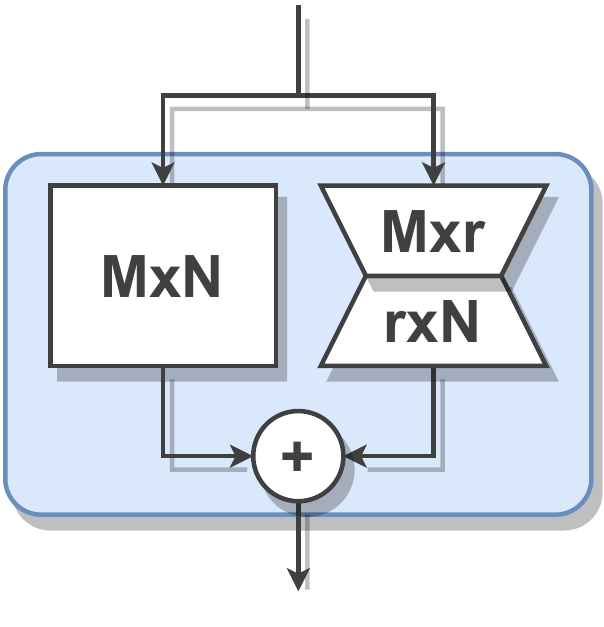} 
   \caption{TA-LoRA: Task-Agnostic Low-Rank Adaptation Module in MTLoRA}
    \label{fig:TALoRA}
   \includegraphics[width=0.9\linewidth]{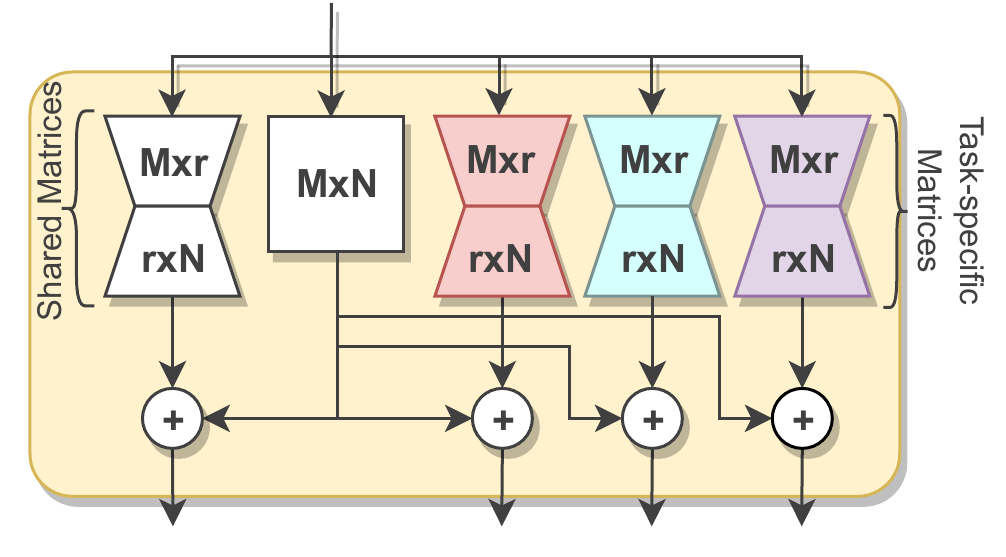}
   \caption{TS-LoRA: Task-Specific Low-Rank Adaptation Module in MTLoRA}
    \label{fig:TSLoRA}
  \end{subfigure}

  \caption{\textit{MTLoRA} framework overview. Task-Agnostic LoRA modules (\textit{TA-LoRA}) are placed at each transformer block, excluding the last ones in each stage where our Task-Specific LoRA (\textit{TS-LoRA}) modules are placed to capture task-specific fine-tuning at different scales.}
  \label{fig:mtlora_overview}
  \vspace{-10pt}
\end{figure*}

\subsection{Low-Rank Adaptation for MTL Architectures}
\label{mtlora}

Low-rank decomposition modules are increasingly used to adapt pre-trained models for various tasks \cite{hu2021lora}. These modules, incorporated into layers that involve matrix multiplication, are notably used in the attention layers of transformer-based models. The function of these modules can be mathematically described as follows:
\begin{equation}
\label{lora}
    Output_{Layer_i} = W_ix + b_i + \alpha B_iA_ix
\end{equation}
Here, $W_i$ and $b_i$ are the original weights and biases of the layer.
$A$ and $B$ are the rank decomposition matrices, and $x$ is the input to the layer $i$.
$\alpha$ is the adaptation scale which controls the deviation of the tuned model from the original model.
During training, only parameters of $A$, $B$, and potentially $b_i$ are trained, significantly reducing the memory footprint and leading to faster training. 
During inference, the $A$ and $B$ matrices can be merged into $W_i$, ensuring that the introduction of low-rank decomposition does not add any extra latency to the inference process.
For Hierarchical Vision transformers, several locations within the architecture are identified as suitable for the application of low-rank matrices, enhancing task adaptability as shown in Figure \ref{fig:backboneMtlora}.
\begin{itemize}
    \item \textbf{QKV Computation in Attention Layers}: The Query, Key, Value (QKV) computation, the main components of the attention mechanism, represents a prime candidate for low-rank adaptation. 
    Fine-tuning this computation allows for modifications to the attention mechanism, making it more suited for specific downstream tasks, which can improve the model's ability to process visual inputs in a task-specific manner.
    \item \textbf{Projection Layer}: The projection layer in transformers is responsible for projecting the attention layer's output back to the original feature space. 
    Fine-tuning the projection layer allows the attention output to be projected into the task's feature space, yielding better performance on downstream tasks.
    \item \textbf{Feed Forward Layers in the MLP block}: These layers, consisting of two dense layers (FC1 and FC2) with nonlinear activation in-between, transform the attention output into the final feature representation. 
    Fine-tuning these layers dictates the model's capacity to effectively generate task-specific final feature representations to be processed by subsequent stages or task-specific decoders.
\end{itemize}

Adopting low-rank decomposition modules in these layers provides controllable knobs to trade-off fine-tuning efficiency (i.e., Number of trained parameters) and adaptation quality (i.e., Performance on the downstream tasks).
We consider two variants of low-rank decomposition modules as shown in Figure \ref{fig:mtlora_overview}: (1) \textit{Task-Agnostic Low-Rank Adaptation} module to capture shared features among the various tasks, and (2) \textit{Task-Specific Low-Rank Adaptation} module that can learn task-specific features.

\noindent \textbf{Task-Agnostic Low-Rank Adaptation (\textit{TA-LoRA}):}
In our framework, we utilize task-agnostic low-rank adaptation modules, which employ low-rank decomposition to adjust the corresponding weights, as detailed in Equation \ref{lora}. 
The \textit{TA-LoRA} modules are designed to identify and leverage shared features across multiple downstream tasks, thereby facilitating knowledge sharing. 
We have integrated \textit{TA-LoRA} modules into the transformer blocks of the Hierarchical Vision Transformer backbone, with the exception of the final block in each stage, as illustrated in Figure \ref{fig:backboneMtlora}. 
Specifically, these \textit{TA-LoRA} modules are applied to adapt key computational layers within the transformer blocks, namely the QKV Layer, the Projection Layer, and the MLP block.
The inclusion of \textit{TA-LoRA} modules aims to promote a balanced and synergistic learning process. 
This approach ensures fine-tuning that is unbiased towards any specific task, preventing overfitting. 
Additionally, to address the challenge of conflicting gradients in MTLoRA, the final block of each stage incorporates our novel \textit{Task-Specific Low-Rank Adaptation} modules, which aim to capture task-specific features as explained in the following paragraph.

\noindent \textbf{Task-Specific Low-Rank Adaptation (\textit{TS-LoRA}):}
One of the main challenges in multi-task low-rank adaptation is to disentangle the feature learning space in order to solve the conflicts between the various downstream tasks.
To achieve that, we propose our novel \textit{TS-LoRA} modules.
\textit{TS-LoRA} incorporates separate task-specific low-rank matrices in addition to the shared low-rank matrices as shown in Figure \ref{fig:TSLoRA}.
These modules are designed to operate in two distinct modes. First, when a \textit{TS-LoRA} module follows a layer with a \textit{TA-LoRA} module (for instance, in the projection layer), it processes the shared input to derive task-specific representations.
Conversely, in scenarios where a layer with a \textit{TS-LoRA} module succeeds another with a similar module (as observed in MLP feed-forward layers), it processes task-specific inputs to produce corresponding task-specific outputs as follows:
\begin{equation}
    Output_{layer_i/task_j} = W_ix + b_i + \alpha_i B_{task_j} A_{task_j}x
\label{mtlora_update}
\end{equation}
Here, $Output_{layer_i/task_j}$ represents the $task_j$'s specialized output at layer $i$. 
$W_i$ and $b_i$ are the original weights and biases of the layer. 
$x$ is input to $layer_i$. 
$B_{task_j}$ and $A_{task_j}$ are the \textit{TS-LoRA} matrices for task $j$.
These modules fine-tune the model according to the specific needs of each task.
The outputs of the \textit{TS-LoRA} modules are directed toward the corresponding task-specific fusion modules and decoders, as shown in Figure \ref{fig:mtlora_overview}.
This allows the encoder to generate task-specific feature representations at various scales.
Since these \textit{TS-LoRA} matrices are only connected to their corresponding task-specific decoders in the computation graph, the backward propagation only updates those matrices according to their corresponding task's loss.

In MTLoRA, the usage of both \textit{TA-LoRA} and \textit{TS-LoRA} modules is key to achieving an optimal balance between generalization and specialization within the MTL model. 
The \textit{TA-LoRA} modules are designed to capture generalized information throughout the model, ensuring that a fundamental level of generality is maintained across various tasks.
In contrast, the \textit{TS-LoRA} modules are used to encapsulate unique updates that are tailored to each specific task. 
This dual-module approach ensures that while the model efficiently processes shared features relevant across multiple tasks, it also possesses the capacity to cater to the specific demands of individual tasks.

\subsection{Multi-Scale Task-Specific Feature Sharing in Encoder-Decoder MTL Architecture}
\label{multi_scale_feature_sharing}
Multi-scale feature propagation within the encoder-decoder architecture has been shown to enhance performance in vision tasks \cite{zhou2019multi,vandenhende2020mti} where the input data is captured at various scales, providing various levels of abstractions.
Typically, a hierarchical vision transformer processes input through multiple stages, with each stage generating features at a different scale. 
Merging features from these different scales results in a more comprehensive feature representation.
In conventional setups, features at various scales are often fused together to create a unified, shared multi-scale feature set applicable to all tasks. 
However, our \textit{TS-LoRA} module allows for a unique specialization at each scale for every task since it generates task-specific features at the end of every transformer stage as shown in Figure ~\ref{fig:backboneMtlora}.
This enables the creation of task-specific multi-scale features which pushes the model to fine-tune the features at each scale according to the requirements of each task. 
Our learnable task-specific multi-scale fusion layers use a residual blocks-based architecture to combine the features at different scales (i.e., receptive fields) in an informative way for every downstream task.

\subsection{Fine-tuning Non-Attention Modules}
\label{training}
Several studies in the domain of parameter-efficient training have highlighted the benefits of unfreezing some low training-cost modules \cite{gao2023llama}, such as layer normalization, which can positively impact the model's performance without significantly increasing the number of trainable parameters. Hence, we explore the effect of unfreezing different modules within MTLoRA.
In addition to training the shared \textit{TA-LoRA} and the task-specific \textit{TS-LoRA} modules, we unfreeze the patch embedding layer, the patch merging layer, the layer normalization, and the position bias in the attention layer.
We provide insights about the effect of freezing each of those layers on the accuracy-efficiency trade-off in Subsection \ref{ablation}.
Additionally, we explore adding low-rank decomposition modules to the \textit{patch merging} module instead of completely unfreezing it.
This allows for further reduction in training parameters; we denote this lighter version as \textit{MTLoRA+} referring to those extra low-rank decomposition modules added outside the transformer blocks.

\section{\textbf{Experimental Results}}
\label{sec:expr}
\subsection{Implementation Details}
\noindent \textbf{Dataset:} 
We evaluate our method on the PASCAL MTL dataset \cite{pascal}. Following other papers in MTL literature \cite{ye2022inverted, vandenhende2020mti, xu2018pad}, we use the PASCAL-Context split that has annotations for various dense prediction tasks such as semantic segmentation, human part detection, surface normals estimation, and saliency distillation.
It has 4,998 images in the training split and 5,105 in the validation split.

\noindent \textbf{Evaluation metrics:}
Following common multi-task learning evaluation practices \cite{vandenhende2020mti}, the semantic segmentation, saliency estimation, and human part segmentation tasks are evaluated using mean intersection over union (mIoU). 
We use the root mean square error (rmse) in the predicted angles to evaluate the surface normals task. 
We also measure the overall performance $\Delta m$ as the average per-task reduction in performance compared to the single-task baseline $st$:
\vspace{-5pt}
\begin{equation}
    \Delta m = \frac{1}{T} \sum_{i=1}^T (-1)^{l_i} (M_i - M_{st,i})/M_{st,i}
\label{overall_accuracy}
\vspace{-5pt}
\end{equation}
where $l_i$ = 1 if a lower value means better for performance measure $M_i$ of task $i$, and 0 otherwise.
The single-task performance is measured for a fully converged model that uses the same backbone network only for that task.

\noindent \textbf{Implementation:} 
\textit{MTLoRA} is implemented using PyTorch, and the code is publicly available on GitHub. Our main artifact is an easily pluggable \textit{MTLoRALinear} layer that encapsulates our \textit{TS-LoRA} and \textit{TA-LoRA} modules, enabling the model to adapt to different tasks by using task-specific low-rank matrices. We use rank 4 for the task-specific matrices while we explore different ranks for the shared matrices.
We adopt the publicly available \textit{Swin Transformer} backbone \cite{liu2021swin}, which was pre-trained on the ImageNet dataset \cite{deng2009imagenet} as our shared encoder.
Then, we attach simple task-specific decoders for different dense tasks.
Specifically, we use a simple decoder similar to the one in HR-Net \cite{hr_net}, which includes linear and bilinear upsampling layers to efficiently perform dense vision tasks, and we adapt the number of output dimensions to different tasks.
The number of decoder parameters is only $6\%$ of the overall MTL model's parameters when using \textit{Swin-Tiny} as a backbone. We run each experiment on a single NVIDIA V100 GPU.

\noindent \textbf{Training:}
To train our multi-task learning model, we use a loss function equal to the weighted sum of the losses of the various downstream tasks as follows:
\vspace{-5pt}
\begin{equation}
  Loss_{MTL} = \sum_{i}^T \omega_{task\_i} \times L_{task\_i}
  \label{eq:mtl_loss}
\vspace{-5pt}
\end{equation} 
\noindent where $\omega_{task\_i}$ and $L_{task\_i}$ are the weight and the loss of the various tasks in the MTL model, respectively.
Specifically, we use the standard per-pixel cross-entropy for semantic segmentation and human part segmentation, $L1$ loss for surface normals estimation, and balanced cross-entropy for saliency detection.
We also adopt the task weights used by Vandenhende \textit{et. al.} \cite{vandenhende2020mti}.

\begin{table*}[t]
  \centering
  \setlength{\tabcolsep}{4.5pt}
\renewcommand{\arraystretch}{0.85}
  \caption{Results \-- MTLoRA versus SOTA parameter efficient training methods. The table summarizes the number of trainable parameters in each method.
  It also includes the accuracy of the downstream tasks as well as the average MTL model's accuracy ($\Delta m$).
  The last column indicates whether or not the model allows all the tasks to be executed simultaneously.
  The symbols $\uparrow$ and $\downarrow$ indicate higher and lower is better, respectively. \textbf{bold} numbers highlights how \textit{MTLoRA} dominates the full finetuning while training $3.6\times$ less parameters.
  }
  \vspace{-5pt}
\small
  \begin{tabular}{c | c  c c c | c | c | c }
    \toprule
    \multirow{2}{*}{\textbf{Method}} & \textbf{SemSeg} & \textbf{Human Parts} & \textbf{Saliency} & \textbf{Normals} & 
      \multirow{2}{*}{$\Delta m (\%) $} & \textbf{Trainable} & \textbf{Single Inference}   \\
      & ($mIoU \uparrow$) & ($mIoU \uparrow$) & ($mIoU \uparrow$) & ($rmse \downarrow$) & & 
      \textbf{Parameters} (M) &
      \textbf{For All Tasks}   \\
    \midrule
    Single Task & 67.21 & 61.93 & 62.35 & 17.97 & 0 & 112.62   & $\times$ \\
    MTL - Tuning Decoders Only & 65.09 & 53.48 & 57.46 & 20.69 & -9.95  & 1.94  & \checkmark  \\
    MTL - Full Fine Tuning  &  67.56 & 60.24 & 65.21 & 16.64 & +2.23 &  30.06    & \checkmark  \\
    \midrule
    Adapter \cite{he2021towards} &  69.21 &  57.38 & 61.28 & 18.83  & -2.71 & 11.24  & $\times$ \\
    Bitfit \cite{zaken2021bitfit} & 68.57 & 55.99 & 60.64 & 19.42 & -4.60 & 2.85   & $\times$  \\
    VPT-shallow \cite{jia2022visual}  & 62.96 &  52.27 & 58.31 & 20.90 & -11.18  & 2.57  & $\times$  \\
    VPT-deep \cite{jia2022visual}  & 64.35 & 52.54 & 58.15 & 21.07 & -10.85 & 3.43  & $\times$ \\
    Compactor \cite{karimi2021compacter} & 68.08 & 56.41 & 60.08 & 19.22 & -4.55 & 2.78   & $\times$  \\
    Compactor++ \cite{karimi2021compacter}  & 67.26 & 55.69 & 59.47 & 19.54 & -5.84 & 2.66    & $\times$ \\
    LoRA \cite{hu2021lora} & 70.12 & 57.73 & 61.90 & 18.96 & -2.17 & 2.87   & $\times$ \\
    VL-Adapter \cite{sung2022vl} & 70.21 & 59.15 & 62.29 & 19.26 & -1.83  & 4.74  & $\times$ \\
    HyperFormer \cite{mahabadi2021parameter}  & 71.43 & 60.73 & 65.54 &  17.77 & +2.64  & 72.77  & $\times$ \\
    Polyhistor \cite{liu2022polyhistor} & 70.87  & 59.15  & 65.54 &  17.77 & +2.34  & 8.96  & $\times$ \\
    \midrule
    MTLoRA ($r=16$) & 68.19 & 58.99 & 64.48 & 17.03  & +1.35 & 4.95 & \checkmark  \\
    MTLoRA ($r=32$) & 67.74 & 59.46 & 64.90 & 16.59 & +2.16 & 6.08  & \checkmark  \\
    MTLoRA ($r=64$) & 67.9 & 59.84 & 65.40 & 16.60 & \textbf{+2.55} & \textbf{8.34}  & \textbf{\checkmark} \\
    \midrule
    MTLoRA+ ($r=4$)  & 68.12 & 57.77 & 63.14 & 17.60 & -0.52 & 2.57 & \checkmark \\
    MTLoRA+ ($r=8$)  & 68.54 & 58.30 & 63.57 & 17.41 & +0.29 & 3.15  & \checkmark \\
    MTLoRA+ ($r=16$) & 68.28 & 58.70 & 64.323 & 17.034 & +1.19  & 4.29 & \checkmark  \\
    \bottomrule
  \end{tabular}
  \label{tab:results}
  \vspace{-10pt}
\end{table*}

\subsection{Baselines}
To evaluate the performance of \textit{MTLoRA}, we compare its accuracy and the number of training parameters to other parameter-efficient training methods.
We first compare to \textit{Single task} baselines where each task has a separate model and all parameters are fine-tuned to minimize the loss on the corresponding task. 
We also build a multi-task learning model with a shared encoder and task-specific decoders and evaluate the post-training accuracy when only decoders are fine-tuned (\textit{MTL - Tuning Decoders Only}) and when the full model is fine-tuned (\textit{MTL - Full Fine-Tuning}) to reduce the overall tasks losses as shown in equation \ref{eq:mtl_loss}.

As mentioned earlier, \textit{MTLoRA} is the first to achieve parameter-efficient training for multi-task learning models. Therefore, to evaluate our method against state-of-the-art methods, we compare \textit{MTLoRA} to other single-task parameter-efficient training methods where a task-wise module is added for each task using the setup provided by Liu \textit{et al.} \cite{liu2022polyhistor}.
Specifically, we compare to the following baselines:
(1) Adapter \cite{he2021towards} where task-specific bottleneck modules are placed into transformer layers
(2) Bitfit \cite{zaken2021bitfit} where only biases, patch merging layers, and patch projection layers are fine-tuned.
(3) VPT \cite{jia2022visual} where tunable embeddings (i.e., 50 embeddings per layer) are inserted in the first input layer (VPT-shallow) and all layers (VPT-deep).
(4) Compacter \cite{karimi2021compacter}, which decomposes the fast matrix into two low-rank vectors, and Compacter++, which only places modules after MLP layers.
(5) LoRA \cite{hu2021lora}, where the low-rank decomposition is applied on attention layers with rank r = 4 and the adapter output scale (i.e., 4), which matches our \textit{MTLoRA} hyperparameter.
(6) VL-Adapter \cite{sung2022vl}, which shares an adapter across different tasks.
(7) Hyperformer \cite{mahabadi2021parameter}, where a hyper-network is used to produce the weights for adapters for various tasks.
(8) Polyhistor \cite{liu2022polyhistor}, where decomposed hyper-networks are used to share information across different tasks while still necessitating separate training and inference paths for each task.

\subsection{Quantitative Analysis}
For a fair comparison, \textit{MTLoRA}, \textit{MTLoRA+}, as well as all other baselines, are all based on the \textit{Swin-Tiny} variant of the Swin Transformers family of models \cite{liu2021swin}.
Table \ref{tab:results} shows the per-task accuracy, the overall MTL accuracy based on Equation \ref{overall_accuracy}, the number of trainable parameters of \textit{MTLoRA} and \textit{MTLoRA+} compared to our baselines.
The last column indicates whether or not the corresponding parameter-efficient training method allows all the tasks to be executed simultaneously.
As mentioned earlier in Figure ~\ref{fig:mtltypes}, the ability to execute all tasks in a single inference path is essential for applications where efficiency and latency are critical.
As shown in Figure \ref{fig:mtlora_vs_sota}, \textit{MTLoRA} and \textit{MTLoRA+} offer a Pareto for the trade-off between the number of trainable parameters and the accuracy of the downstream tasks.

\subsection{Ablation}
\label{ablation}
\begin{figure}[t]
  \centering
\vspace{-10pt}
\includegraphics[width=\linewidth]{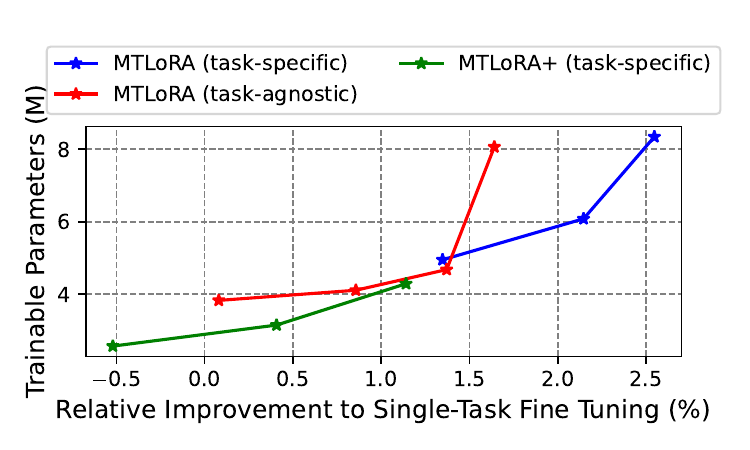}
\vspace{-20pt}
\caption{Accuracy versus trainable parameters of MTLoRA with task-agnostic vs task-specific adaptation modules.}
\label{fig:ranks}
\end{figure}

\noindent \textbf{Effect of task-specific modules in MTLoRA:}
To show the effectiveness of the task-specific Low-rank-decomposition modules in \textit{MTLoRA}, we compare the performance of \textit{MTLoRA} and \textit{MTLoRA+} to a similar setup with only task-agnostic low-rank decomposition modules.
The results of this comparison, as shown in Figure \ref{fig:ranks}, clearly demonstrate the impact of adding task-specific adaptation modules. 
Notably, the integration of these modules results in a substantial improvement in the accuracy-efficiency trade-off during parameter-efficient fine-tuning.
This enhancement indicates the ability of the task-specific modules to effectively untangle the parameter space involved in MTL. 
Consequently, this leads to positive knowledge sharing during the fine-tuning process, significantly boosting the performance of each downstream task.

\noindent \textbf{Effect of Various Backbone Adaptation Locations:}
As mentioned in Subsection \ref{mtlora}, we insert low-rank decomposition modules in four different locations in the Hierarchical Transformer encoder: 1) the feed-forward layers in the MLP block (FC1 and FC2), 2) the QKV layer, and 3) the projection layer.
We analyze the effect of removing the low-rank decomposition modules from each of those layers to get insights about the effectiveness of adapting each of those weights.
Table \ref{tab:lora_locations} shows the overall MTL accuracy ($\Delta m$) versus the number of trainable parameters when the low-rank decomposition modules are removed from each of the four locations.
Those results are from MTLoRA applied on a Swin-Tiny backbone where all low-rank decomposition modules have a rank of 32.
We can see that adapting the QKV weights is the most important since removing its low-rank modules causes the most accuracy degradation.
On the other hand, removing the adaptation from the first linear layer of the MLP block seems the least effective in comparison.
However, the table shows that each low-rank decomposition module offers a significant improvement in the overall performance of the multiple downstream tasks.
Removing some of them can be used to achieve a different accuracy efficiency trade-off during training.

\begin{table}[t]
  \centering
  \setlength{\tabcolsep}{2.2pt}
\renewcommand{\arraystretch}{0.85}
  \caption{Effect of removing the various low-rank decomposition matrices in \textit{MTLoRA} from the different locations in the backbone vision transformer.
  \textit{None} refers to the default of MTLoRA, where all the low-rank modules are adopted.}
\vspace{-5pt}
\small
  \begin{tabular}{c | c | c | c | c | c}
    \toprule
    & \textit{None} & \textit{No FC1} & \textit{No FC2} & \textit{No Proj} & \textit{No QKV} \\
    \midrule
    $\Delta m$ & +2.55 & +1.93\% & +1.65\% & +1.65\% & +1.26\% \\
    Trainable (M) & 8.34 &  6.81 &  6.81 & 7.73 & 7.21 \\
    \bottomrule
  \end{tabular}
  \vspace{-10pt}
  \label{tab:lora_locations}
\end{table}

\noindent \textbf{Effect of Freezing Non-Attention Modules:}
As mentioned earlier in SubSection \ref{training}, besides fine-tuning the low-rank decomposition matrices, we unfreeze the patch embedding layer, the patch merging layer, the layer normalization, and the position bias in the attention layer.
We analyze the effect of freezing these extra modules to provide insights into the accuracy-efficiency trade-off associated with each component.
Table \ref{tab:freeze_modules} shows the overall MTL accuracy ($\Delta m$) versus the number of trainable parameters when each of those modules is frozen.
We can notice that unfreezing all those modules yields the highest accuracy.

\begin{table}[t]
  \centering
  \setlength{\tabcolsep}{2.2pt}
\renewcommand{\arraystretch}{0.85}
  \caption{Effect of freezing the different modules outside the transformer block.}
\vspace{-5pt}
\small
  \begin{tabular}{c | c | c | c | c}
    \toprule
    & \textit{Patch} & \textit{Layer} & \textit{Position} & \textit{Patch} \\
     & \textit{Embed} & \textit{Norm} & \textit{Bias} & \textit{Merging} \\
    \midrule
    Accuracy (W/ ST) & +2.02 & +2.06 & +1.99 & +1.74 \\
    Training Parameters (M) & 8.34 & 8.32 & 8.32 & 6.80 \\
    \bottomrule
  \end{tabular}
  \label{tab:freeze_modules}
  \vspace{-10pt}
\end{table}

\begin{figure}[t]
  \centering
\includegraphics[width=0.95\linewidth]{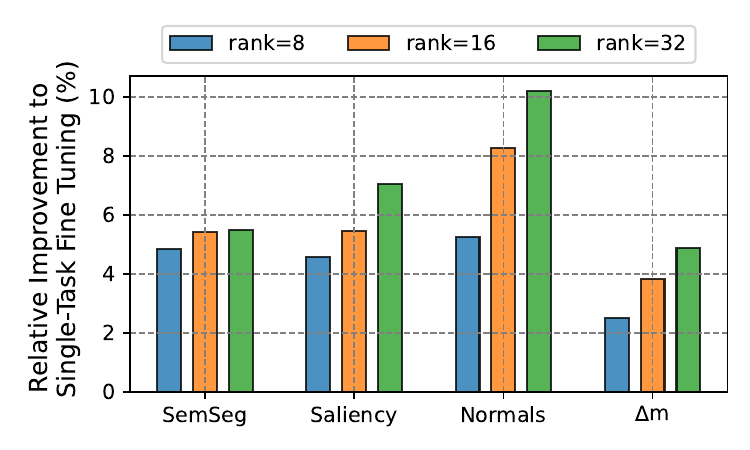}
\vspace{-5pt}
\caption{Performance of \textit{MTLoRA} on various downstream tasks when applied to a \textit{Swin-Base} model pre-trained on \textit{ImageNet-22K}}
\label{fig:scales_backbones}
\vspace{-15pt}
\end{figure}

\noindent \textbf{Results with Larger Backbones and Pre-training Datasets:}
To analyze the efficacy of \textit{MTLoRA} scaled backbone and pre-training datasets. 
We apply \textit{MTLoRA} on \textit{Swin-Base} that was pre-trained on \textit{ImageNet22k} dataset.
Figure \ref{fig:scales_backbones} shows the improvement in the accuracy of \textit{MTLoRA} compared to single-task models. 
We can see that MTLoRA scales well with a larger backbone, providing significant improvement compared to the single-task models while training significantly fewer parameters.
More analyses are included in the Appendix.



\section{\textbf{Conclusion}}
\label{sec:conclusion}
In conclusion, this work introduced \textit{MTLoRA}, an innovative framework designed to enable parameter-efficient training for Multi-Task Learning (MTL) models. Central to \textit{MTLoRA} are the Task-Agnostic and Task-Specific Low-Rank Adaptation modules, which are instrumental in effectively disentangling the parameter space during MTL fine-tuning. 
Those modules allow the fine-tuning to balance both task specialization and interaction within MTL environments. 
We have demonstrated the application of \textit{MTLoRA} in hierarchical-transformer-based MTL architectures, tailoring them to a variety of downstream dense prediction tasks. 
Our experiments show that \textit{MTLoRA} not only surpasses the accuracy of fully fine-tuned MTL models but also achieves this with a substantially lower number of trained parameters ($3.6\times$ reduction in trainable parameters). Additionally, \textit{MTLoRA} provides Pareto-optimality in the trade-off between the number of trainable parameters and accuracy compared to existing state-of-the-art parameter-efficient training approaches.

{
    \small
    \bibliographystyle{ieeenat_fullname}
    \bibliography{refbib}
}

\clearpage
\setcounter{page}{1}
\maketitlesupplementary

\begin{table*}[h]
  \centering
  \setlength{\tabcolsep}{4pt}
  \renewcommand{\arraystretch}{1}
    \small
  \caption{\textit{MTLoRA} versus SoTA parameter-efficient training methods when applied on Pyramid Vision Transformer \cite{pvt} backbone. The table summarizes the number of trainable parameters in each method.
  It also includes the accuracy of the downstream tasks as well as the average MTL model's accuracy ($\Delta m$).
  }
  \begin{tabular}{c | c c c c | c | c}
    \multirow{2}{*}{\textbf{Method}} & \textbf{SemSeg} & \textbf{Human Parts} & \textbf{Saliency} & \textbf{Normals} & 
      \multirow{2}{*}{$\Delta m (\%) $} & \textbf{Trainable Parameters}   \\
      & ($mIoU \uparrow$) & ($mIoU \uparrow$) & ($mIoU \uparrow$) & ($rmse \downarrow$) & & (M)  \\
    \midrule
    Single Task & 68.81 & 61.27 & 62.67 & 17.55 & 0.00 &  97.51  \\
    MTL - Fine Tune Decoders & 64.86 & 51.18 & 61.54 & 19.55 & -8.85 & 2.11 \\
    \midrule
    Compactor++ \cite{karimi2021compacter}  & 70.29 & 54.80 & 63.16 & 18.82 & -3.71 & 2.20  \\
    BitFit \cite{zaken2021bitfit} & 71.41 & 55.71 & 64.08 & 18.69 & -2.38 & 2.34 \\
    LoRA \cite{hu2021lora} & 71.89 & 56.9 & 64.27 & 18.48 & -1.35 & 2.41 \\
    Adapter \cite{he2021towards} & 71.94 &  56.38 & 64.16 & 18.75  & -1.97 & 2.90  \\
    Polyhistor \cite{liu2022polyhistor} & 71.00 & 57.52 & 65.83 & 17.83 & +0.13 & 7.32 \\ 
    Hyperformer \cite{mahabadi2021parameter} & 70.81 & 57.76 & 65.49 & 17.75 & +0.14 & 16.14 \\
    \midrule
    \textit{MTLoRA ($r = 64$)} & 69.74 & 58.08 & 65.62 & 17.35 & \textbf{+1.2} & 8.69 \\
    \bottomrule
  \end{tabular}
  \label{tab:diff_vit}
\end{table*}

\begin{table*}[t]
  \centering
  \setlength{\tabcolsep}{4pt}
  \renewcommand{\arraystretch}{1}
    \small
  \caption{Evaluating MTLoRA with $rank = 32$ when applied on Swin-Tiny backbone pretrained on ImageNet-22K dataset and various decoders for the downstream dense prediction tasks.
  The table summarizes the number of parameters for each decoder as well as the total number of parameters.
  It also includes the accuracy of the downstream tasks as well as the average MTL model's accuracy ($\Delta m$).
  }
  \begin{tabular}{c | c c c c | c | c}
    \multirow{2}{*}{\textbf{Decoder}} & \textbf{SemSeg} & \textbf{Human Parts} & \textbf{Saliency} & \textbf{Normals} & 
      \multirow{2}{*}{$\Delta m (\%) $} & \textbf{Trainable Parameters} (M)   \\
      & ($mIoU \uparrow$) & ($mIoU \uparrow$) & ($mIoU \uparrow$) & ($rmse \downarrow$) & & 
      $Decoder/All$  \\
    \midrule
    HR-Net \cite{hr_net} & 69.44 & 61.08 & 63.24 & 16.47 & +2.93 & 1.94 $/$ 6.08 \\
    SegFormer \cite{xie2021segformer} & 69.59 & 61.13 & 63.74 & 16.62 & +3.00 & 2.08 $/$ 6.22 \\
    ASPP \cite{aspp} & 72.32 & 60.98 & 63.04 & 16.51 & +3.83 & 12.44 $/$ 16.58 \\
    \bottomrule
  \end{tabular}
  \label{tab:different_decoders}
\end{table*}

\section{Different pretraining datasets: ImageNet-1K vs ImageNet-22K}
\begin{figure}[t]
  \centering
\includegraphics[width=0.95\linewidth]{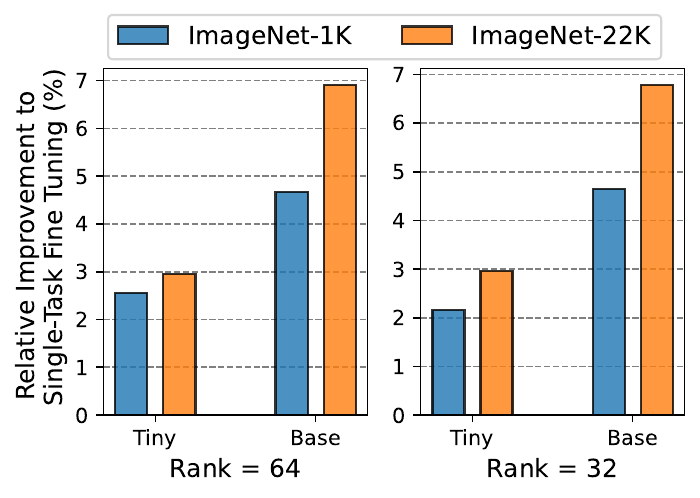}
\vspace{-10pt}
\caption{Performance with MTLoRA with different pretraining datasets on various Swin-Transformer backbones.}
\label{fig:22k_vs_1k}
\vspace{-15pt}
\end{figure}

We analyze the effect of using different pretraining datasets on the performance of MTLoRA.
Figure \ref{fig:22k_vs_1k} shows the relative improvement in accuracy compared to the single task models of MTLoRA when applied to Swin-Tiny and Swin-Base pretrained on ImageNet-$1K$ and ImageNet-$22K$.
The figure shows that the pretraining dataset can have a significant impact on the model's performance.
Specifically, using a model pre-trained on a richer dataset (i.e., ImageNet-22K) results in a better performance on the downstream tasks without any additional cost on our parameter-efficient training methodology.

\section{MTLoRA with different Adaptation Scales}
\begin{figure}[t]
  \centering
\includegraphics[width=0.95\linewidth]{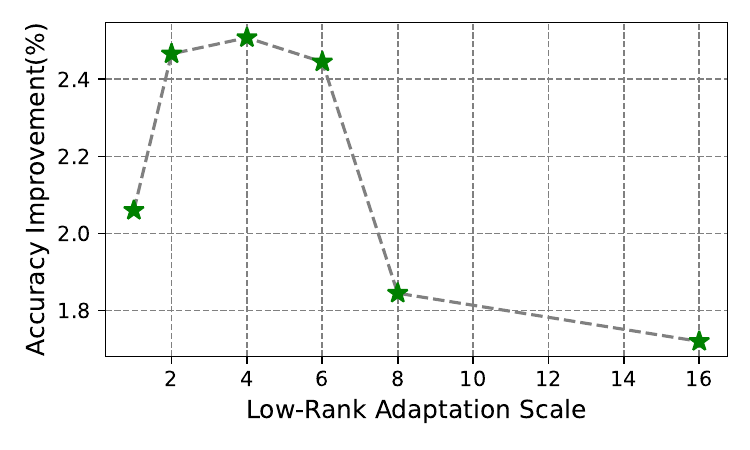}
\vspace{-10pt}
\caption{Effect of the hyper-parameter $\alpha$ on the accuracy of MTLoRA on the downstream tasks.}
\label{fig:scales}
\vspace{-10pt}
\end{figure}

The scale value $\alpha$ in Equation \textcolor{red}{2} determines how much the fine-tuned model can deviate from the original baseline model.
For example, A scale value of 0 is the same as not using the LoRA weights and only using the base model weights, and a scale value of 1 means that both the LoRA weights as well as the base model weights have the same influence. 
It is common to use $\alpha$ in ${1, 2}$ for language models \cite{hu2021lora}; however, we experiment with different scales to analyze their effect when fine-tuning vision transformers for multiple tasks.
Figure \ref{fig:scales} shows the overall accuracy at different scales.
We found that empirically, scale $\mathbf{4}$ performs the best for the purpose of MTLoRA.

\section{MTLoRA with different Backbones}
To ensure the generalizability of MTLoRA, we apply it to another vision transformer backbone \-- Pyramid Vision Transformer \cite{pvt}.
We analyze the impact of using MTLoRA when applied on \textit{PVT-Small}.
Table ~\ref{tab:diff_vit} shows that applying MTLoRA offers Pareto-optimal accuracy-efficiency trade-off compared to state-of-the-art paramter-efficient training techniques.
Specifically, MTLoRA achieves higher accuracy ($\Delta m$) when compared to Hyperformer \cite{mahabadi2021parameter} while training 2$\times$ fewer parameters.

\section{MTLoRA with different decoders}
We analyze the effect of using different decoders on the performance of MTLoRA.
We choose 3 commonly used decoders for dense prediction tasks: (1) HRNet \cite{hr_net}, which interpolates and concatenates multi-scale features from the hierarchical backbone and then passes them to a couple of MLP layers. (2) SegFormer \cite{xie2021segformer}, which uses MLP layers to combine the multi-scale features from the hierarchical backbone, then parse them to one last MLP layer for prediction. (3) Atrous Spatial Pyramid Pooling (ASPP) \cite{aspp}, which has a more complicated architecture utilizing spatial pyramid pooling capable of extracting multi-scale contextual information by
probing the incoming features with filters or pooling operations at multiple rates and multiple effective fields-of-view.
We plug those decoders into a pretrained Swin-Tiny backbone, then use our MTLoRA technique to train the model to perform multiple downstream tasks.
Table \ref{tab:different_decoders} shows that MTLoRA generalizes perfectly when different decoders are used.
Different decoders can provide different accuracy-efficiency trade-offs, which provide flexibility to adapt the training budget depending on the application requirements and the available resources.

\section{MTLoRA with different number of tasks}
In this section, we evaluate MTLoRA in a setting with increased number of tasks.
Table \ref{tab:num_tasks} demonstrates the accuracy-efficiency trade-off with increasing task numbers. The table shows that integrating additional tasks into MTLoRA incurs minimal latency (in terms of trainable parameters) relative to single-task learning and conventional full MTL fine-tuning, while still maintaining superior accuracy compared to these approaches.

\begin{table}[t]
  \centering
  \setlength{\tabcolsep}{5pt}
  \renewcommand{\arraystretch}{0.8}
    \small
  \caption{Analyzing the number of FLOPs for different numbers of tasks for \textit{Individual Task-Specific Adaptation} versus our \textit{Shared Multi-Task Adaptation} depicted in Figures 2(a) and 2(b) respectively. We can clearly notice that our approach is significantly more efficient as the number of tasks increase.}
   \vspace{-7pt}
  \begin{tabular}{c | c | c }
    & \textbf{Individual Task-} & \textbf{Shared Multi-}  \\
    & \textbf{Specific Adaptation} & \textbf{Task Adaptation} \\
    & (Figure 2(a)) & (Figure 2(b)) \\
    \midrule
    1 Tasks (GFLOPs) & 18.48 & 18.48 \\
    2 Tasks (GFLOPs) & 36.91 & 19.50 \\
    3 Tasks (GFLOPs) & 55.32 & 20.49 \\
    4 Tasks (GFLOPs) & 73.74 & 21.49 \\

    \bottomrule
  \end{tabular}
  \label{tab:flops}
  \vspace{-5pt}
\end{table}


\section{FLOPs overhead for different number of tasks.} 
Thank you for the suggestion. As you mentioned, while adding tasks incurs a slight overhead in backbone inference, this cost is marginal as it avoids divergent paths unlike previous techniques. Moreover, these tasks can be batched for efficiency, similar to SWIN's window-based computations, due to their low-rank nature. We've included Table ~\ref{tab:flops} to compare FLOPs against task numbers in \textit{Individual Task-Specific Adaptation} and our \textit{Shared Multi-Task Adaptation} (Figures 2(a) and 2(b)). This highlights the minimal impact on FLOPs in our approach compared to traditional \textit{Individual Task-Specific Adaptation}.


\begin{table}[t]
  \centering
  \setlength{\tabcolsep}{1pt}
  \renewcommand{\arraystretch}{0.8}
    \small
  \caption{Evaluating MTLoRA for \textbf{different numbers of tasks}.}
  \vspace{-5pt}
  \begin{tabular}{c | c | c}
    \multirow{2}{*}{\textbf{Method}} & \multirow{2}{*}{\textbf{$\Delta m (\%) $}} & \textbf{Trainable}   \\
      & & \textbf{Parameters} (M)  \\
    \midrule
    Single Task (All 4 tasks) & 0 & 112.62 \\
    Full MTL Fine-Tuning (All 4 tasks) & +2.23 &  30.06   \\
    \midrule
    MTLoRA (SemSeg and Normals) & +8.7 & 5.83 \\
    MTLoRA (SemSeg and Sal) & +5.2 &  5.83 \\
    MTLoRA (Semseg, Normals, and Sal) & +4.37 &  6.45 \\
    MTLoRA (All 4 tasks) & +2.55 & 8.34 \\
    \bottomrule
  \end{tabular}

  \label{tab:num_tasks}
  \vspace{-5pt}
\end{table}

\end{document}